\documentclass{article}
\usepackage{spconf,amsmath,graphicx}
\usepackage{pgfplots}
\usepackage{stfloats}
\pgfplotsset{compat=newest}

\title{EFFICIENT CONVOLUTIONAL NEURAL NETWORK FOR AUDIO EVENT DETECTION}

\name{Matthias Meyer, Lukas Cavigelli, Lothar Thiele}
\address{Computer Engineering and Networks Laboratory, ETH Zurich, Switzerland\\
\{matthias.meyer, thiele\}@tik.ee.ethz.ch, cavigelli@iis.ee.ethz.ch}

\begin{document}

\maketitle
\begin{abstract}
Wireless distributed systems as used in sensor networks, Internet-of-Things and cyber-physical systems, impose high requirements on resource efficiency. Advanced preprocessing and classification of data at the network edge can help to decrease the communication demand and to reduce the amount of data to be processed centrally. In the area of distributed acoustic sensing, the combination of algorithms with a high classification rate and resource-constraint embedded systems is essential. Unfortunately, algorithms for acoustic event detection have a high memory and computational demand and are not suited for execution at the network edge. This paper addresses these aspects by applying structural optimizations to a convolutional neural network for audio event detection to reduce the memory requirement by a factor of more than 500 and the computational effort by a factor of 2.1 while performing 9.2\,\% better.
\end{abstract}
\begin{keywords}
Acoustic Event Detection, Convolutional Neural Networks, Low-Power Embedded Systems, Acoustic Sensor Networks, Mobile Computing
\end{keywords}
\section{Introduction} 
\label{sec:intro}

Many applications for sensor networks, cyber-physical system or Internet-of-Things require a low power consumption for long-term autonomous operation. Local preprocessing and classifying data at the edge nodes can be a solution to reduce data transmission and therefore, to reduce energy consumption. In addition, such an approach can avoid that enormous amounts of data need to be communicated to and processed by a centralized data analysis infrastructure.

The detection of acoustic events is a typical example: Instead of streaming audio through the network to perform server-side event detection, the acoustic events of interest can be pre-detected directly at the sensing device reducing the network's data throughput significantly.

The accurate detection and classification of individual acoustic events from a sound-emitting environment is of interest for many application such as surveillance \cite{cristani_audio-visual_2007} or environmental monitoring \cite{girard_custom_2012}. The low-power embedded devices used for such systems, however, come with very stringent memory and throughput limitations. These resource constraints impose severe limits to the complexity of suitable event detection algorithms.

Recently, a convolutional neural network (CNN)-based approach has been proposed for acoustic event detection \cite{takahashi_deep_2016} using a network design adapted from image classification \cite{simonyan_very_2014}. It has been shown that this approach outperforms previous state-of-the-art approaches by a large margin. Unfortunately, 
CNNs are computationally expensive, and the proposed algorithm also comes at the expense of having a huge memory requirement due to the huge number of parameters. For image classification a way to reduce the complexity of CNN-based classification systems has been presented \cite{springenberg_striving_2014}. These two approaches are joined in this paper to build a highly-accurate CNN capable of running on embedded platforms with limited resources.

In this context, the present paper contains the following contributions:
\begin{itemize}
\item
A novel algorithm for acoustic event detection is presented that focuses on improving the accuracy while reducing the memory requirements and number of operations.

\item
The algorithm is experimentally verified and compared to a state-of-the-art convolutional neural networks for acoustic event detection. 
The experiment shows that the overall reduction of memory requirement by a factor of 515 and a reduction of operations by a factor of 2.1 does not affect performance and the accuracy even increases by 9.2\,\%.
\end{itemize}

\section{Related Work}
\label{sec:relwork}
 
Hardware platforms for battery-operated devices have stringent power requirements. On such low-power devices which are limited to a few 100\,mW, like the ARM Cortex M7 series, the overall on-chip storage is typically limited to less than 2\,MB and the available digital signal processing performance is limited to a few 100 millions multiply-accumulate operations per second even on the most advanced components \cite{nxp_kv5x,st_stm32f7}. 

To achieve maximum energy efficiency while using CNNs, System-on-Chips (SoCs) with hardware accelerators for CNN workload or more generally 2D convolutions can be considered. Such platforms can provide speed-ups by a factor of around $100\times$ and an improved energy efficiency of about $40\times$ \cite{conti2015ultra,cav2015origami}. Such system perform optimal if the CNN comprises a simple and structured architecture. While this concept can provide a relief on the admissible computational effort, the strong limitations on available memory remain because any external memory would deteriorate the device's energy efficiency substantially. By removing memory-demanding, non-convolutional layers a CNN architecture is ideal to maximize the efficiency of hardware convolution accelerators \cite{rossi_pulp:_2015} 

For acoustic event detection different algorithms have been presented based on Non-Negative Matrix Factorization \cite{komatsu_acoustic_2016}, Hidden Markov Models \cite{zhou_hmm-based_2008} or Recurrent Neural Networks \cite{parascandolo_recurrent_2016}. Like in many other machine learning applications, CNNs have been proven to be the key for high classification accuracy. The advantage of such an architecture for acoustic event detection is its inherent inclusion of a temporal neighborhood since acoustic events are strongly characterized by temporal changes. Besides the mentioned CNN for acoustic event detection, so far CNNs have been used mainly for speech recognition \cite{abdel-hamid_convolutional_2014} or music \cite{dieleman_end--end_2014} classification tasks. These algorithms are all computationally expensive, and the current state-of-the-art also comes at the expense of having a huge number of parameters \cite{takahashi_deep_2016}. Implementing such a network on mobile devices or sensor nodes is difficult due to memory and computational restrictions on these devices. A way to reduce the complexity of CNN-based classification systems has been presented for image classification \cite{springenberg_striving_2014} and a similar structure is used in \cite{choi_automatic_2016}. 

These approaches are joined in this paper to build a highly-accurate CNN for acoustic event detection capable of running on embedded platforms with limited resources.

\section{Model Architecture}
\label{sec:architecture}
\begin{figure}[htb]
\begin{minipage}[b]{1.0\linewidth}
  \centering
  \centerline{\includegraphics[width=8.5cm]{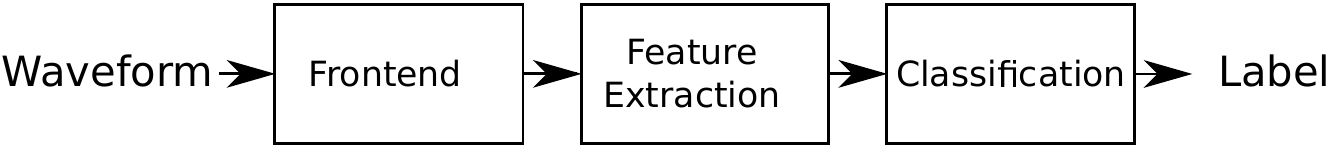}}
\end{minipage}
\caption{Model architecture}
\label{fig:model}
\end{figure}

CNNs, by default, are not designed to run on low-power embedded devices, thus a careful design in terms of structure and learning algorithms must be chosen.
To explain the different steps which are necessary to detect an acoustic event the detection system is divided into three major components, which are illustrated in Figure ~\ref{fig:model}. First, the raw time-domain audio waveform is transformed by a front-end into a time-frequency representation. Then the systems extracts features from this representation. In a final step the features are classified.

In the following, for each step the best option in terms of the challenges mentioned in section \ref{sec:relwork} is highlighted and chosen. 

\subsection{Front-end}
\label{sec:frontend}
The front-end is used to transform the raw audio signal into a time-frequency representation from which features in both, frequency- and time-domain, can be extracted. In general these transforms are based on the Short-Time Fourier Transform (STFT), which can be efficiently implemented via the Fast Fourier Transformation (FFT). This fact makes it favorable compared to novel techniques \cite{dieleman_end--end_2014}, which show good results by learning filterbank coefficients from the audio data, but have the major drawback of employing large filterbanks.

Mel-scaling, which mirrors the human auditory system, is often used \cite{cakir_polyphonic_2015} as addition to the STFT in order to compensate for its linear frequency scale. These additional processing steps may also reduce the amount of data that needs to be processed in later stages of the processing pipeline, which is important since the number of MAC operations of a CNN is directly related to the size of the input field. 

As a consequence, in this work a mel-spectrogram has been chosen as front-end to satisfy the real-time constraint. It is calculated with a window size of 32\,ms and a hop size of 10\,ms using a hamming window. The number of mel coefficients is 64. From the spectrogram multiple frames consisting of 400 vectors are extracted. These frames are input to the feature extractor, thus the network analyzes a time span of 4\,s for each frame. 

\begin{figure}[htb]
	\begin{minipage}[b]{1.0\linewidth}
		\centering
		\centerline{\includegraphics[width=8.5cm]{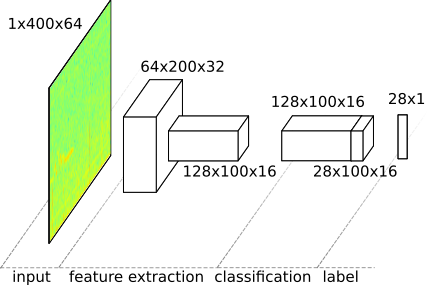}}
	\end{minipage}
	\caption{Data flow of proposed CNN}
	\label{fig:dataflow}
\end{figure}

\begin{table*}[bp]
    \begin{tabular}{ccc|ccc|ccc}
        \multicolumn{3}{c}{CNN-FC} & \multicolumn{3}{c}{CNN-C} & \multicolumn{3}{c}{CNN-CNP} \\
        
        Layer type & \# param. & \# MAC & Layer type & \# param. & \# MAC & Layer type & \# param. & \# MAC \\ 
        \hline 
        
        input & 0 & 0 & 
        input & 0 & 0 & 
        input & 0 & 0 \\
        
        frontend & 25.6\,k & 12.7\,M & 
        frontend & 25.6\,k & 12.7\,M & 
        frontend & 25.6\,k & 12.7\,M \\ 
        
        conv 3, 1, 64 & 1.8\,k & 32.3\,M & 
        conv 3, 1, 64 & 640 & 14.8\,M & 
        conv 3, 1, 64 & 640 & 14.8\,M \\
        
        conv 3, 1, 64 & 36.9\,k & 656.9\,M & 
        conv 3, 1, 64 & 36.9\,k & 943.7\,M & 
        conv 3, 2, 64 & 36.9\,k & 236.0\,M \\
        
        max pool 1x2 & 0 & 0 & 
        max pool 2x2 & 0 & 0 & 
        - & - & - \\
        
        conv 3, 1, 128 & 73.9\,k & 581.0\,M & 
        conv 3, 1, 128 & 73.9\,k & 471.9\,M & 
        conv 3, 1, 128 & 73.9\,k & 471.9\,M \\
        
        conv 3, 1, 128 & 147.6\,k & 1040.5\,M & 
        conv 3, 1, 128 & 147.6\,k & 943.7\,M & 
        conv 3, 2, 128 & 147.6\,k & 236.0\,M\\
        
        max pool 2x2 & 0 & 0 & 
        max pool 2x2 & 0 & 0 & 
        - & - & - \\ \hline
        
        fc 1024        & 231.2\,M & 231.2\,M & 
        conv 3, 1, 128 & 147.6\,k & 236.0\,M & 
        conv 3, 1, 128 & 147.6\,k & 236.0\,M \\
        
        fc 1024        & 1.1\,M  & 1.1\,M & 
        conv 1, 1, 128 & 16.5\,k & 26.2\,M & 
        conv 1, 1, 128 & 16.5\,k & 26.2\,M \\
        
        fc 28         & 28.7\,k & 28.7\,k & 
        conv 1, 1, 28 & 3.6\,k  & 5.7\,M & 
        conv 1, 1, 28 & 3.6\,k  & 5.7\,M\\
        
        - & - & - & 
        avg pool & 0 & 0 & 
        avg pool & 0 & 0 \\
        
        activation & 0 & 0 & 
        activation & 0 & 0 & 
        activation & 0 &  0 \\
        
        \hline
        Total: & 233\,M & 2555\,M & 
        Total: & 452\,k & 2655\,M & 
        Total: & 452\,k & 1239\,M \\
    \end{tabular} 
	\caption{Structure, number of parameters and number of required MAC operations for three CNNs. The first (CNN-FC) uses fully-connected layers as classifier, the second (CNN-C) uses convolutional layers as classifier, the third (CNN-CNP) uses convolutional layers as classifier but no max pooling layers. Convolutional layers are defined as \textit{conv filter\_size, stride, number\_of\_filters}. Fully-connected layers as \textit{fc output\_dimensions}. Max pooling layers as \textit{max pool pool\_size} }
	\label{table:params}
\end{table*}

\subsection{Feature Extraction}
\label{sec:featureextraction}
The feature extraction block uses a CNN to learn the features of the time-frequency representation. Most CNNs are built from very few basic building blocks: convolution, activation and pooling layers. The concatenation of these blocks introduces a higher depth to the network which has been shown to enhance accuracy \cite{simonyan_very_2014}. 
A higher depth results in a higher number of required operations and a higher parameter count. In this work the feature extraction block is therefore limited to two sections consisting of two convolutional layers each, which provides enough parameters to learn significant features but is still moderate in parameter count. Moreover, one convolutional layer with a higher filter size (e.g. 5x5) can be reduced with layers using 3x3 filters, which reduces the number parameters and potentially even improves the classification performance \cite{simonyan_very_2014}. In most CNNs max pooling is used to regularize the network, but is has been shown that for small scale datasets the removal of the max pooling layer does not affect the performance \cite{springenberg_striving_2014}. As a consequence max pooling is removed from the network and the stride of the preceding layer is increased by 1, which divides the required MAC operations for this layer by four while maintaining the same network structure in principle.

\subsection{Classification}
Table \ref{table:params} illustrates the structure of a network with fully-connected dense layers as classifier (CNN-FC) and a network using only convolutional layers as classifiers (CNN-C). For both networks the number of parameters and the number of required MAC operations are listed. It becomes obvious that the fully-connected layers have the biggest impact on parameter count which makes the preceding convolutional layers almost negligible. It has been shown that a fully-connected layer can be replaced by a 1x1 convolutional layer \cite{lin_network_2013}, which reduces the number of parameters, and thus the memory footprint, from $233\times10^{6}$ to $452\times10^{3}$. The reduction of parameters also further regularizes the network, which is an advantage for training the network and therefore, an improvement in accuracy can be expected.
As a last layer average pooling reduces the output of the last convolutional layer to an array, whose size matches the number of labels. The replacement of the fully-connected layers slightly increases the number of MAC operations.

\subsection{Final design}
The design proposed in this paper is denoted as CNN-CNP and is illustrated in Table \ref{table:params}. After applying each optimization step as described above the parameter count is decreased considerably by a factor of 515 to 452\,k parameters and the number of MAC operations by a factor of 2.1 to 1239\,M\,MAC. Moreover, after optimization the network consists only of convolutional layers. This unified architecture beneficial for hardware implementation and especially for the use of convolutional hardware accelerators.

\section{Experiment}
After applying these fundamental changes to the network and substantially reducing the number of parameters and arithmetic operations, it needs to be validated that the classification performance is maintained and kept on an acceptable level. For this purpose the two networks (CNN-C and CNN-CNP) have been implemented using Keras \cite{chollet2015keras}. These networks are compared against the best performing implementations from \cite{takahashi_deep_2016}, which are referred to as A and B. The network A has the same structure as the CNN-FC network from Table \ref{table:params}, the B network is a more complex network with a higher depth and bigger fully-connected layers. 

\subsection{Dataset}
\label{sec:dataset}
The dataset \cite{takahashi_deep_2016} contains various sound files collected from freely available online sources. It consists of 28 different event types of variable length, e.g. airplanes, violins, birds or footsteps. The total length of all 5223 audio files is 768.4 minutes. The data is split into training and test set. The training set contains 75\,\% of the original data and is further subdivided into training and validation set with a ratio of 0.25. 
Although the dataset is strongly biased no data augmentation was performed in the following experiments, since the main focus is on the comparison of the algorithm in terms of structure, resources and classification performance, and not on the augmentation technique.
Both networks were trained by minimizing the cross-entropy loss using the gradient-based optimizer ADAM \cite{kingma_adam:_2014} with mini-batches of size 128. The optimizers' parameters were left at its default values.

Testing was done by predicting the probabilities for each class on a 4 seconds window randomly extracted from the test set. The class with the highest probability was chosen as the correct class and compared against the ground truth.

\subsection{Results}
\label{sec:results}
The experimental results are listed in Table \ref{table:accuracy}. The values for accuracy of network A and B are taken from the original publication, the values for the number of parameters and number of MACs are calculated based on the information taken from the original publication.
The first line of  Table \ref{table:accuracy} shows the accuracy results for networks A and B with data augmentation which are 91.7\,\% and 92.8\,\%, respectively. As expected, these are better than the corresponding accuracy results without complex data augmentation which are 77.9\,\% and 80.3\,\%, respectively. 
The networks CNN-C and CNN-CNP have an accuracy of 86\,\% and 85.1\,\%, respectively. Thus, without sophisticated data augmentation both proposed networks perform better than the reference network A and even better than the more complex network B.

The analysis of parameter count shows that when 16\,bit parameters are assumed, the total memory consumption for the CNN-FC network's weights is 466\,MB, which is not feasible for most edge computing applications considering the fierce power and memory constraints for distributed sensing devices. In contrast, the  weight storage for the CNN-C and CNN-CNP is approx. 904\,kB. Even when considering additional overhead by the implementation on low-power devices such as the NXP KV5x or ST STM32F7 \cite{nxp_kv5x,st_stm32f7}, their flash memory of up to 2\,MB is still sufficient to store the parameters of the presented acoustic event detection algorithm. 

Considering that the devices mentioned above have a processing capability higher than 430\,M\,MAC/s and processing the input buffer is only required every 4 seconds, they are able to handle the 1239\,M\,MACs of CNN-CNP in less than 4 seconds. Thus the classification can be considered real-time. In practical applications, however, the claimed processing capability might not be reached or the necessary real-time performance is higher. Therefore the use of hardware accelerators is suggested, allowing to offload the computation to a dedicated processor while maintaining a good energy efficiency. As has been explained, the novel network structure matches the communication and memory access pattern that is expected by current accelerators.

The evaluation shows that the design specifications could be reduced considerably without impact on performance.

\begin{table}
    \begin{tabular}{c|cccc}
    	Model & A, \cite{takahashi_deep_2016} & B, \cite{takahashi_deep_2016}& CNN-C & CNN-CNP  \\
    	\hline
    	Accuracy & 91.7\,\% & 92.8\,\%  & -  & - \\
    	w/ aug & & & & \\
    	Accuracy  & 77.9\,\% & 80.3\,\% & 86.0\,\% & 85.1\,\%\\
    	w/o aug & & & & \\
    	\# params & 233\,M & 257\,M & 452\,k & 452\,k \\
    	\# MACS & 2543\,M & 3533\,M & 2655\,M & 1239\,M  \\
    	\hline
    \end{tabular}
	\caption{Accuracy (with and without data augmentation), parameter count and number of operations for the proposed networks CNN-C and CNN-CNP compared to the top scoring implementations A and B from \cite{takahashi_deep_2016}.}
	\label{table:accuracy}
\end{table}

\section{Conclusion}
\label{sec:conclusion}
In this paper, an acoustic event detection algorithm was presented that exploits the advantages of CNNs while being implementable on low-power microcontrollers. First, a convolutional neural network has been proposed that is able to supersede state-of-the-art acoustic event detection algorithms. Second, the network can be efficiently implemented on  resource-limited low-power embedded devices.
It was demonstrated that it is possible to reduce the memory requirement by a factor of 515 and the number of operations by a factor of 2.1, while outperforming a similar network with fully-connected layers by 9.2\,\%.
The  structured approach of the CNN consisting mainly of convolutional layers makes it easily portable to novel convolutional hardware accelerators which can further increase the energy efficiency.

\section{Acknowledgement}
The work presented in this paper was scientifically evaluated by the SNSF, and financed by the Swiss Confederation and by nano-tera.ch.
\vfill\pagebreak

\bibliographystyle{ieeetr} 
\bibliography{refs}

\end{document}